%% file: main.tex
\documentclass{article}

\usepackage[final, nonatbib]{neurips_2019}

\usepackage[utf8]{inputenc} 
\usepackage[T1]{fontenc}    
\usepackage{hyperref}       
\usepackage{url}            
\usepackage{booktabs}       
\usepackage{amsfonts}       
\usepackage{nicefrac}       
\usepackage{microtype}      
\usepackage{xcolor}
\usepackage{graphicx}
\usepackage{subcaption}
\usepackage{algorithm}
\usepackage{algorithmic}

\usepackage{amsmath}
\usepackage{amssymb}

\usepackage[normalem]{ulem}
\usepackage[square, numbers]{natbib}
\title{VIABLE: Fast Adaptation via Backpropagating Learned Loss}

\usepackage[symbol]{footmisc}

\author{
  Leo Feng\thanks{Correspondence to: leo.feng@keble.ox.ac.uk} \\
  University of Oxford\\
  \And
  Luisa Zintgraf \\
  University of Oxford \\
  \And
  Bei Peng \\
  University of Oxford \\
  \And
  Shimon Whiteson \\
  University of Oxford \\
  Latent Logic \\
}

\begin{document}

\maketitle

\input{core/introduction.tex}
\input{core/background.tex}

\input{core/contribution.tex}
\input{core/experiments.tex}
\input{core/discussion.tex}

\newpage

\subsubsection*{Acknowledgements}
We thank Andrei Rusu for useful feedback on working with the LEO image embeddings \cite{rusu2019meta}. This work was supported by a generous equipment grant from NVIDIA. Luisa Zintgraf is supported by the Microsoft Research PhD Scholarship Program. This project has received funding from the European Research Council (ERC) under the European Union’s Horizon 2020 research and innovation programme (grant agreement number 637713).

\bibliography{main}

\newpage

\input{core/appendix.tex}

\end{document}

%% file: core/introduction.tex
\section{Introduction}

Meta-learning is a popular and general way to tackle few-shot learning problems, i.e., learning how to solve unseen tasks given only little data.
Many meta-learning methods can be characterised as meta-gradient-based \cite{finn2017model, li2017meta, rusu2019meta, zintgraf2019fast}.
Briefly speaking, meta-gradient-based methods work as follows.
During training, at each iteration, these methods perform a gradient-based task-specific update (often referred to as the "inner loop"). 
Then, for the meta-update, so-called \textit{meta-gradients} are computed by backpropagating through these inner loop updates (which therefore involves taking higher order gradients).
At test time, on a new task, only the inner-loop update is performed using a few gradient updates.
In few-shot learning, typically, the loss function applied at test time is the one we are ultimately interested in minimising, such as the mean-squared-error loss for a regression problem. 
However, given we have few samples at test time, we argue that the loss function we want to minimise is not necessarily the loss function most suitable for computing gradients in a few-shot setting. 
Such a loss function is naive in the sense that it treats each datapoint independently, disregarding any relationships between them.
This can be particularly problematic when only few datapoints are given and include, e.g., outliers or correlated points. 
Furthermore, it can be prone to cause over- or underfitting \cite{mishra2017simple}, depending on the stepsize and number of gradient steps.
Therefore, we propose to instead \textit{learn} the test-time loss function for meta-gradient-based methods for few-shot adaptation.
In this work, we introduce \textit{fast adaptation \textbf{via} \textbf{b}ackprogating \textbf{le}arned loss} (VIABLE), a generic meta-learning extension which builds on existing meta-gradient-based methods by learning a differentiable loss function using meta-gradients.
This loss function replaces the pre-defined inner-loop loss function and is meta-learned such that it maximises performance (i.e., minimises the pre-defined loss) within a few gradient steps and with little data. 
We show that learning a loss function capable of leveraging relational information between samples reduces underfitting, and significantly improves performance and sample efficiency on a simple regression task. 
In addition, we show VIABLE is scalable by evaluating on the Mini-Imagenet dataset \citep{ravi2016optimization}.
Since we typically use neural networks as function approximators, we will refer to the network making predictions as the \textit{prediction network} and the learned loss function as the \textit{loss network}.

Learning a loss function has been explored in a variety of ways in machine learning fields \cite{andrychowicz2016learning, chebotar2019metalearning, duan2016rl, houthooft2018evolved, santos2017learning, sung2017learning, veeriah2019discovery, wang1611learning, wu2018learning} including reinforcement learning and semi-supervised learning. In this paper, we are concerned with the few-shot supervised learning setting.
Closest related to our method is recent work by \citet{chebotar2019metalearning}, who propose $ML^3$, in which they learn a loss function in a similar fashion as VIABLE.
In contrast to our work, $ML^3$ is not designed for few-shot learning and instead uses the learned loss function to learn a prediction network \textit{from scratch} per task. 
VIABLE on the other hand can be applied on top of any meta-gradient-based meta-learning techniques designed for few-shot learning.
Also closely related is work by \citet{sung2017learning}, who propose meta-critics. 
In addition to also learning \textit{from scratch} per task, during meta-training, the meta-critic (loss network) is updated after each batch of task-specific actor (prediction network) updates;
while in VIABLE, the loss network is frozen during task-specific updates and thus requires far fewer updates in total.
Most importantly, compared to the above methods, we propose to learn a loss function that is designed to operate on \textit{the entire dataset} at once, thus leveraging relational information between datapoints.
We achieve this by using a relation network \cite{santoro2017simple} that looks at pairwise combinations of datapoints. As we show in this paper, this leads to a significant improvement in terms of performance.

%% file: core/background.tex
\section{Background}

We consider the problem setting of meta-learning for supervised learning problems. In supervised learning, we learn a model $f:x\mapsto \hat{y}$ that maps data points $x$ that have a true label $y$ to predictions $\hat{y}$. 
In few-shot learning problems, during each meta-training iteration, a batch of $N$ tasks $\mathbf{T} = \{\mathcal{T}_i\}_{i=1}^N$ is sampled from a task distribution $p(\mathcal{T})$.
A task $\mathcal{T}_i$ is a tuple ($\mathcal{X}$, $\mathcal{Y}$, $\mathcal{L}$, $q$), where $\mathcal{X}$ is the input space, $\mathcal{Y}$ is the output space, $\mathcal{L}$ is the task-specific loss function, and $q(x, y)$ is a distribution over data points. 
During each meta-training iteration, for each $\mathcal{T}_{i} \in \mathbf{T}$, we sample from $q_{\mathcal{T}_{i}}$: $\mathcal{D}_i^\text{train} = \{ (x, y)^{i,m}\}_{m=1}^{M_i^\text{train}}$ and $\mathcal{D}_i^\text{test} = \{ (x, y)^{i, m}\}_{m=1}^{M_i^\text{test}}$, where $M^{train}_i$ and $M^{test}_{i}$ are the fixed number of training and test datapoints respectively. 
The training data is used to perform updates on the model $f$. Afterwards, the updates are evaluated on the test data and $f$ or the update rule are adjusted.

\subsection{Context Adaptation via Meta-Learning: CAVIA}

In theory, VIABLE can be generically applied to meta-gradient-based methods.
In this paper, we evaluate on CAVIA \cite{zintgraf2019fast} because it applies the inner-loop update only on a small set of so-called context parameters instead of the entire network, making it easier to optimise.
CAVIA aims to learn two distinct sets of parameters: task-specific context parameters $\phi$ and task-agnostic parameters $\theta$. 
At every meta-training iteration (inner loop), CAVIA starts from a fixed value $\phi_0$, typically $\phi_0 = 0$, and updates its context-parameters $\phi$ for each task $\mathcal{T}_i$ in the current batch $\mathbf{T}$ of tasks as follows\footnote{We outline CAVIA for one gradient update step, but it can be extended to several gradient steps.}:
\begin{equation} \label{eq:cavia_inner_update}
	\phi_i = 
	\phi_0 - 
	\alpha 
	\nabla_{\phi}
	\frac{1}{M_i^\text{train}}
	\sum\limits_{(x, y) \in \mathcal{D}_i^\text{train}} 
	\mathcal{L}_{\mathcal{T}_i} (f_{\phi_0, \theta}(x), y)
\end{equation}
In the meta-update step (outer loop), the model parameters $\theta$ are updated with respect to the performance after the inner-loop update:
\begin{equation} \label{eq:cavia_outer_update}
	\theta \leftarrow \theta - 
	\beta
	\nabla_\theta 
	\frac{1}{N}
	\sum\limits_{\mathcal{T}_i \in \mathbf{T}}
	\frac{1}{M_i^\text{test}}
	\sum
	\limits_{(x, y) \in \mathcal{D}_i^\text{test}}
	\mathcal{L}_{\mathcal{T}_i} (f_{\phi_i, \theta}(x), y)
\end{equation}
At test time, model parameters $\theta$ are frozen and only the task-specific parameters $\phi$ are updated.

%% file: core/contribution.tex
\section{Fast Adaptation via Backpropagating Learned Loss: VIABLE}

\begin{figure}
  \centering
  \includegraphics[width=0.62\linewidth]{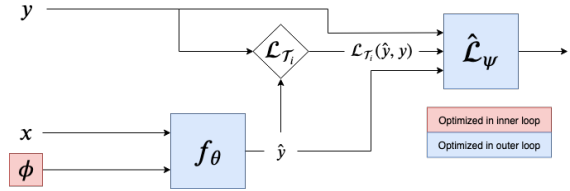}
  \caption{
  Overview of VIABLE with a \textit{simple loss network} applied to CAVIA, where $f_{\theta}$ is the prediction network, $\hat{\mathcal{L}}_{\psi}$ is the loss network, and $\mathcal{L}_{\mathcal{T}_i}$ is the original task-specific loss function. 
  } 
\end{figure}

We introduce VIABLE, a generic meta-learning extension that aims to adapt a loss function applicable to meta-gradient-based methods.
During training, at each iteration, VIABLE trains an existing meta-gradient-based method (referred to as \textit{prediction network}) by performing gradient updates using the output of a differentiable learned loss function (referred to as \textit{loss network}). 
During the meta-update step, the meta-gradients are calculated and used to update the \textit{loss network}.
In this section, we consider two variants of loss networks: a simple loss network and an extension inspired by relation networks \cite{santoro2017simple} which leverages relationships between datapoints.

\textbf{Simple Loss Network.} First, we consider a simple loss network $\hat{\mathcal{L}}_{\psi}$ which takes as input the target $y$, the prediction $\hat{y}$, and pre-defined task-specific loss $\mathcal{L}_{\mathcal{T}_i}(\hat{y}, y)$, and outputs a loss value. 
In the inner loop of the meta-gradient-based method, we replace the pre-defined task-specific loss with the output of our loss network. 
In this case, we replace CAVIA's inner loop update (see \eqref{eq:cavia_inner_update}) with:
\begin{equation}
\label{eq:inner_update_lf}
    \phi_i = 
    \phi_0 - 
    \alpha 
    \nabla_{\phi}
    \frac{1}{M_i^\text{train}}
    \sum\limits_{(x, y) \in \mathcal{D}_i^\text{train}} 
    \hat{\mathcal{L}}_{\psi} (\mathcal{L}_{\mathcal{T}_i}(f_{\phi_0, \theta}(x), y), f_{\phi_0, \theta}(x), y)
\end{equation}
The task-specific parameters $\phi$ are updated by backpropagating the learned loss \textit{through the original loss and the outputs of the prediction network}. In the outer loop, we update the parameters of the loss network $\psi$ along with the task-agnostic parameters of the prediction network $\theta$ (see \eqref{eq:cavia_outer_update}):
\begin{equation} \label{eq:outer_update_lf}
    \psi \leftarrow \psi - 
    \gamma
    \nabla_\psi 
    \frac{1}{N}
    \sum\limits_{\mathcal{T}_i \in \mathbf{T}}
    \frac{1}{M_i^\text{test}}
    \sum\limits_{(x, y) \in \mathcal{D}_i^\text{test}}
    \mathcal{L}_{\mathcal{T}_i} (f_{\phi_i, \theta}(x), y) 
\end{equation}
\textbf{Relation Loss Network.} 
Note that the pre-specified loss function $\mathcal{L}_{\mathcal{T}_i}$ and the aforementioned simple loss network naively calculate an independent loss per sample and average, ignoring any possible relationships between datapoints. 
For example, in the case of an outlier with a large disagreeing gradient compared to the other samples, simply averaging the gradients may negatively impact the model's performance post-update.
In addition, there is substantial evidence in few-shot learning showing that incorporating relational information between samples improves predictions \cite{koch2015siamese, rusu2019meta, sung2018learning, vinyals2016matching}.
Thus, we believe that loss functions can improve upon gradient-based methods by providing the prediction network with relational information between samples, especially in gradient-based methods like MAML which treat their datapoints as independent during prediction.
To show this, we introduce a relation loss network which takes as input the pairwise combinations of $x$, $y$, $\hat{y}$, $\mathcal{L}_{\mathcal{T}_i}(\hat{y}, y)$. 
Thus, we replace CAVIA's inner loop update (see \eqref{eq:cavia_inner_update}) with:
\begin{equation}
    \phi_i = 
    \phi_0 - 
    \alpha 
    \nabla_{\phi}
    \frac{1}{(M_i^\text{train})^{2}}
    \sum\limits_{(x_j, y_j) \in \mathcal{D}_i^\text{train}} 
    \sum\limits_{(x_k, y_k) \in \mathcal{D}_i^\text{train}} 
    \hat{\mathcal{L}}_{\psi} 
        (
        \mathcal{L}_{\mathcal{T}_i}(\hat{y}_j, y_j), 
        x_j,
        \hat{y}_j, 
        y_j, 
        \mathcal{L}_{\mathcal{T}_i}(\hat{y}_k, y_k), 
        x_k,
        \hat{y}_k, 
        y_k
        )
\end{equation}
where $\hat{y}_j = f_{\phi_0, \theta}(x_j)$. 
Similar to the simple loss network, in the outer loop, we update the loss network and the task-agnostic parameters of the prediction network (see \eqref{eq:outer_update_lf} and \eqref{eq:cavia_outer_update}).

%% file: core/experiments.tex
\section{Experiments}
\label{sec:experiments}
In this section, we evaluate the benefits of replacing the existing loss function in meta-gradient-based meta-learning methods with an adapted loss trained with VIABLE. 
We show that: 
1) a loss function that leverages relational information between samples yields a substantial increase in performance over loss functions without relational information,
2) VIABLE improves the sample efficiency and reduces underfitting in a simple regression task, and
3) VIABLE is scalable by evaluating on the Mini-Imagenet dataset.
For these experiments, we denote simVIABLE as applying VIABLE with a simple loss network to CAVIA, and relVIABLE as applying VIABLE with a relation loss network to CAVIA.
Note that we do not evaluate against $ML^3$ since it is not designed for few-shot learning and thus would require more samples.
We describe the specifics of our implementation in the Appendix.

\subsection{Regression}

We begin with a regression problem of fitting sine curves from \citet{finn2017model}.
A task is defined by the amplitude and phase of the sine curve which are uniformly sampled from $[0.1, 0.5]$ and $[0, \pi]$ respectively. 
During training, for each task, $k$ (default $k=10$) datapoints are uniformly sampled from $x\in[-5, 5]$ and given to the model to perform inner loop updates. 
The task specific loss is mean-squared-error (MSE) loss. 
In these experiments, we perform a single inner-loop update.

\textbf{Improved performance.} Both versions of VIABLE significantly outperform CAVIA.  
With 2 context parameters, CAVIA achieves a loss of 0.21, simVIABLE achieves 0.14, and relVIABLE achieves 0.02, which suggests that leveraging relational information between samples can substantially improve the effectiveness of the loss function. See Appendix \ref{appendix:regression-experiments} for the full results.

\textbf{Improved data efficiency.} For this experiment, we uniformly sample $k \in \{0,\dotsc,20\}$ (the number of training sample points) during training.
We observe in Table \ref{table:vary-sample-points-viable} that relVIABLE achieves better performance with 4 sample points than CAVIA does with 20. 
In Figure \ref{fig:sine-wave}, we see that with only a single gradient update, CAVIA underfits on the 4 test points while relVIABLE fits the curve closely.

\begin{minipage}{0.3\linewidth}
    \centering
    \includegraphics[width=\linewidth]{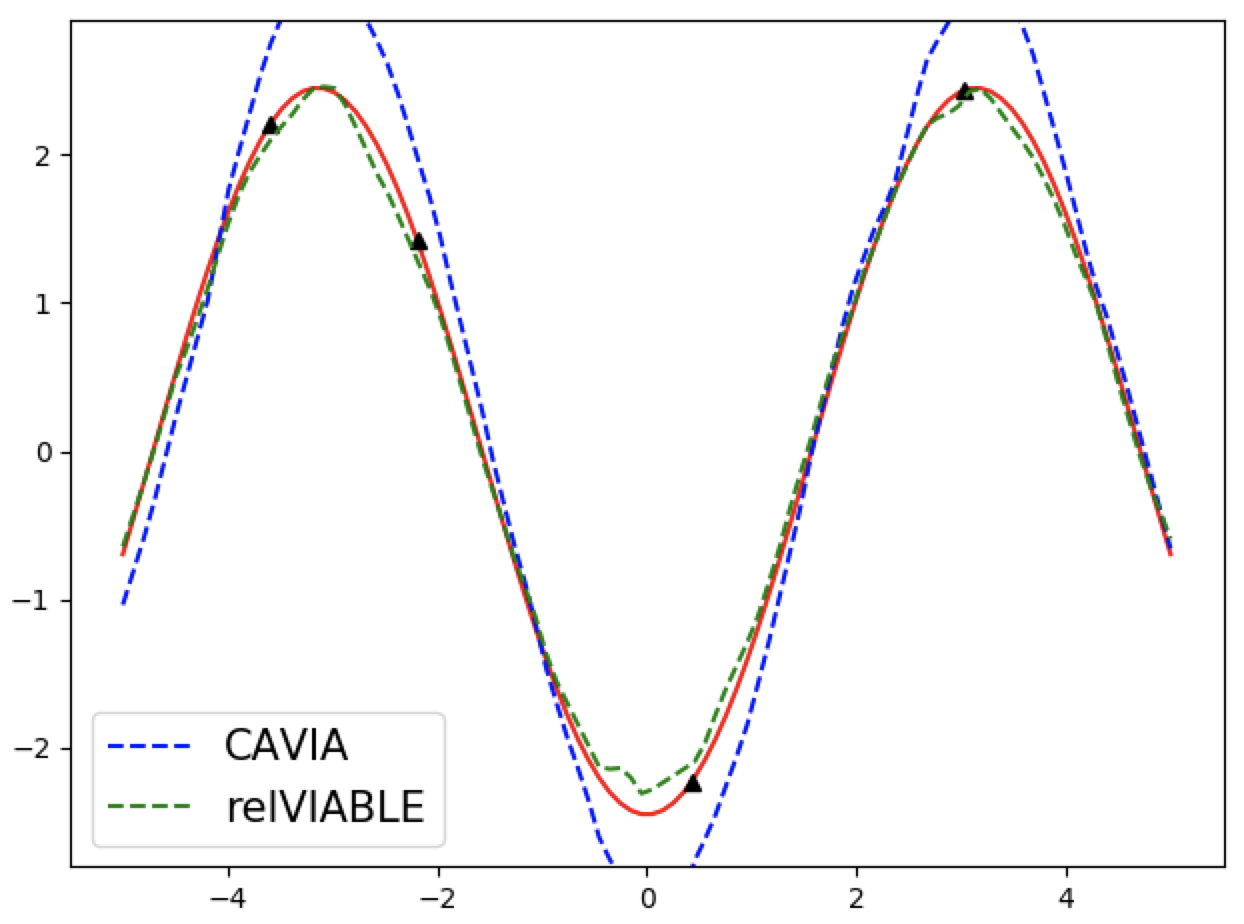}
    \captionof{figure}{Test with 4 data points}
    \label{fig:sine-wave}
\end{minipage}
\hfill
\begin{minipage}{0.65\linewidth}
    \centering
    \begin{tabular}{lcccccc}
      \toprule
                        & \multicolumn{6}{c}{Number of Sample Points}                   \\
      \cmidrule(r){2-7}
      Method            & 0                 & 1 & 2 & 3 &4                 & 20                \\
      \midrule
    CAVIA                   &  $\mathbf{3.13}$     & $1.69$      & $0.93$      & $0.58$       & $0.47 $    & $0.13$       \\
    simVIABLE      &  $\mathbf{3.13}$ & $1.57$ & $0.85$ & $0.45$ & $0.37$ & $0.09$ \\
    relVIABLE      &  $3.14 $     & $\mathbf{1.44}$      & $\mathbf{0.52}$       & $\mathbf{0.17}$      & $\mathbf{0.11}$       & $\mathbf{0.02}$       \\
      \bottomrule
    \end{tabular}
    \captionof{table}{Results for the sine curve regression task. Shown is the MSE for varying number of sample points.}
    \label{table:vary-sample-points-viable}
\end{minipage}\hfill

\subsection{Classification}

We show that this method can scale to problems which require larger networks by testing it on the few-shot image classification benchmark Mini-Imagenet \citep{ravi2016optimization}. 

\textbf{Setup.} 
In \citet{rusu2019meta}, a Wide Residual Network (WRN) \cite{zagoruyko2016wide} is trained with supervised classification on the meta-train set; the network is then frozen and feature representations of the Mini-Imagenet dataset is extracted. 
Following their training protocol, we use the same embeddings and meta-learn on both the meta-train and meta-validation sets, with early-stopping on meta-validation. 

\begin{table}[h]
  \centering
  \begin{tabular}{lcc}
    \toprule
                                        & \multicolumn{2}{c}{5-way accuracy}                   \\
    \cmidrule(r){2-3}
    Method                              & 1-shot                   & 5-shot    \\
    \midrule
    Matching Networks \cite{vinyals2016matching}                   & $46.6\%$                 & $60.0\%$ \\
    MAML \cite{finn2017model}                                & $48.70 \pm 1.84\%$       & $63.11 \pm 0.92\%$ \\
    Meta-SGD* \cite{li2017meta}          & $54.24 \pm 0.03\%$       & $70.86 \pm 0.04\%$ \\
    LEO*  \cite{rusu2019meta}            & $61.76 \pm 0.08\%$       & $77.59 \pm 0.12\%$ \\
    MetaOptNet-SVM-trainval \textsuperscript{$\dagger$} \cite{lee2019meta}     & $\mathbf{64.09 \pm 0.62\%}$       & $\mathbf{80.00 \pm 0.45\%}$ \\
    \midrule
    CAVIA*              & $58.10 \pm 0.51\%$        & $67.07 \pm 0.45\%$ \\
    simVIABLE*          & $57.88 \pm 0.49\%$        & $69.32 \pm 0.41\%$    \\
    relVIABLE*          & $58.26 \pm 0.50\%$        & $70.23 \pm 0.41\%$   \\
    \bottomrule
  \end{tabular}
  \caption{Few-shot classification results on Mini-Imagenet (average accuracy with 95\% confidence intervals). \textsuperscript{$\dagger$} Is the current state-of-the-art. * Used the feature embeddings from \citet{rusu2019meta}}
  \label{table:classification-results}
\end{table}

\textbf{Results.} 
Table \ref{table:classification-results} shows that simVIABLE offers a notable $2.25\%$ improvement over CAVIA while relVIABLE offers a substantial $3.16\%$ increase in accuracy in 5-way 5-shot experiments.
In both variants of VIABLE, 5-way 1-shot experiments are within confidence intervals.
We suspect that learning a loss for 1-shot experiments does not offer a significant advantage due to a single sample being all the information the model is provided regarding a class of images.
For example, there is no concept of an outlier with a single sample.
In the regression experiments, Table \ref{table:vary-sample-points-viable} shows similar results where the learned loss provides minor improvements over CAVIA for a single sample point.

%% file: core/discussion.tex
\section{Conclusion and Future Work}

We proposed VIABLE, a general-purpose meta-learning extension applicable to existing meta-gradient-based meta-learning methods. 
We show that learning a loss capable of leveraging relations between samples through VIABLE improves upon CAVIA by mitigating underfitting and yielding substantial improvements to sample efficiency and performance.
Furthermore, we show VIABLE is scalable by evaluating on the Mini-Imagenet dataset.
For future work, we are interested in applying this extension to other existing meta-learning methods such as MAML and LEO, and evaluating variants of loss networks which utilise more than just pairwise relations such as an attention network.

%% file: core/appendix.tex
\appendix

\begin{center}
{\textbf{VIABLE: Fast Adaptation via Backpropagating Learned Loss}} \\ \vspace{0.3cm}
{\centering \Large Supplementary Material} 
\end{center}

\section{Pseudocode} \label{eq:appendix_code}

\begin{algorithm}[H] \label{alg:simVIABLE}
	\caption{simVIABLE: VIABLE applied to CAVIA with a simple loss network}
	\begin{algorithmic}[1] 
		\REQUIRE Distribution over tasks $p(\mathcal{T})$
		\REQUIRE Step sizes $\alpha$, $\beta$, $\gamma$
		\REQUIRE Initial model $\mathcal{L}_{\psi}$ with $\psi$ intitialised randomly and model $f_{\phi_0, \theta}$ with $\theta$ initialised randomly and $\phi_0=0$ 
		\WHILE{not done}
		\STATE Sample batch of tasks $\mathbf{T} = \{\mathcal{T}_i\}_{i=1}^N$ where $\mathcal{T}_i\sim p$
		\FORALL{$\mathcal{T}_i\in\mathbf{T}$}
		\STATE $\mathcal{D}_i^\text{train}, \mathcal{D}_i^\text{test} \sim q_{\mathcal{T}_i}$
		\STATE $\phi_0 = 0$ 
		\STATE $\phi_i = 
		\phi_0 - 
		\alpha
		\nabla_{\phi}
		\frac{1}{M_i^\text{train}}
		\sum\limits_{(x, y) \in \mathcal{D}_i^\text{train}} 
		\hat{\mathcal{L}}_{\psi} 
		(
		\mathcal{L}_{\mathcal{T}_i}(f_{\phi_0, \theta}(x), y), 
		f_{\phi_0, \theta}(x), 
		y
		)$
		\ENDFOR
		\STATE $	\psi \leftarrow \psi - 
		\gamma
		\nabla_\psi 
		\frac{1}{N}
		\sum\limits_{\mathcal{T}_i \in \mathbf{T}}
		\frac{1}{M_i^\text{test}}
		\sum\limits_{(x, y) \in \mathcal{D}_i^\text{test}}
		\mathcal{L}_{\mathcal{T}_i} (f_{\phi_i, \theta}(x, y))$ 
		\STATE $	\theta \leftarrow \theta - 
		\beta
		\nabla_\theta 
		\frac{1}{N}
		\sum\limits_{\mathcal{T}_i \in \mathbf{T}}
		\frac{1}{M_i^\text{test}}
		\sum\limits_{(x, y) \in \mathcal{D}_i^\text{test}}
		\mathcal{L}_{\mathcal{T}_i} (f_{\phi_i, \theta}(x, y))$ 
		\ENDWHILE
	\end{algorithmic}
\end{algorithm}

\begin{algorithm}[H] \label{alg:relVIABLE}
	\caption{relVIABLE: VIABLE applied to CAVIA with a relation loss network}
	\begin{algorithmic}[1] 
		\REQUIRE Distribution over tasks $p(\mathcal{T})$
		\REQUIRE Step sizes $\alpha$, $\beta$, $\gamma$
		\REQUIRE Initial model $\mathcal{L}_{\psi}$ with $\psi$ intitialised randomly and model $f_{\phi_0, \theta}$ with $\theta$ initialised randomly and $\phi_0=0$ 
		\WHILE{not done}
		\STATE Sample batch of tasks $\mathbf{T} = \{\mathcal{T}_i\}_{i=1}^N$ where $\mathcal{T}_i\sim p$
		\FORALL{$\mathcal{T}_i\in\mathbf{T}$}
		\STATE $\mathcal{D}_i^\text{train}, \mathcal{D}_i^\text{test} \sim q_{\mathcal{T}_i}$
		\STATE $\phi_0 = 0$ 
		\STATE $
		\phi_i = 
		\phi_0 - 
		\alpha
		\nabla_{\phi}
        \frac{1}{(M_i^\text{train})^{2}}
        \sum_{
            \substack{
            (x_j, y_j) \in \mathcal{D}_i^\text{train} \\ 
            (x_k, y_k) \in \mathcal{D}_i^\text{train} 
            }
        }
        \hat{\mathcal{L}}_{\psi} 
            (
            \mathcal{L}_{\mathcal{T}_i}(\hat{y}_j, y_j), 
            x_j,
            \hat{y}_j, 
            y_j, 
            \mathcal{L}_{\mathcal{T}_i}(\hat{y}_k, y_k), 
            x_k,
            \hat{y}_k, 
            y_k
            )$
		
		\ENDFOR
		\STATE $	\psi \leftarrow \psi - 
		\gamma
		\nabla_\psi 
		\frac{1}{N}
		\sum\limits_{\mathcal{T}_i \in \mathbf{T}}
		\frac{1}{M_i^\text{test}}
		\sum\limits_{(x, y) \in \mathcal{D}_i^\text{test}}
		\mathcal{L}_{\mathcal{T}_i} (f_{\phi_i, \theta}(x, y))$ 
		\STATE $	\theta \leftarrow \theta - 
		\beta
		\nabla_\theta 
		\frac{1}{N}
		\sum\limits_{\mathcal{T}_i \in \mathbf{T}}
		\frac{1}{M_i^\text{test}}
		\sum\limits_{(x, y) \in \mathcal{D}_i^\text{test}}
		\mathcal{L}_{\mathcal{T}_i} (f_{\phi_i, \theta}(x, y))$ 
		\ENDWHILE
	\end{algorithmic}
\end{algorithm}

\section{Additional Related Work}

\textbf{Meta-gradient based Methods.} A common form of meta-learning is to adapt parameters in two interleaving phases that can be characterised as the task-specific updates (often referred to as the "inner loop") and the meta-updates (often referred to as the "outer loop"). 
At test time, on a new task, only the task-specific updates are applied. 
\citet{finn2017model} introduces a meta-gradient-based method (MAML) that aims to learn a model initialisation that allows for fast adaptation to a new task given a few task-specific updates. 
Many methods that are inspired by or built on top of MAML can also be classified as meta-gradient-based \cite{antoniou2018train, behl2019alpha, finn2017one, finn2018probabilistic, li2017meta, zintgraf2019fast}.
Another meta-gradient-based method, CAVIA \cite{zintgraf2019fast} extends MAML by splitting the model parameters are into task-specific (context) parameters and task-agnostic parameters, resulting in fewer parameters to optimize in test time.
\citet{rusu2019meta} introduces a meta-gradient-based method LEO that learns to produce network weights from task-specific embeddings. 
In this paper, we focus on CAVIA due to its structure being simple and easy to optimise.

\textbf{Learning a Loss Function.} Specially designed loss functions have been important in improving performance of many tasks such as classification \cite{nguyen2013algorithms}, machine translation \cite{bahdanau2017actor, shen2015minimum}, ranking \cite{taylor2008softrank}, and object detection \cite{song2016training}. 
In recent years, there has been interest in exploring methods for learning a good loss function automatically in a variety of machine learning fields \cite{andrychowicz2016learning, chebotar2019metalearning, duan2016rl, houthooft2018evolved, santos2017learning, sung2017learning, veeriah2019discovery, wang1611learning, wu2018learning}, including reinforcement learning and semi-supervised learning.
In this work, we focus on meta-learning, specifically the few-shot supervised learning setting.
Closely related is meta-critics \cite{sung2017learning} and $ML^3$ \cite{chebotar2019metalearning}, who both learn a form of loss network. In contrast to their works, we are not required to learn our prediction network \textit{from scratch} per task. 
Furthermore, VIABLE is applicable to any meta-gradient-based meta-learning techniques designed for few-shot learning, and, in contrast to meta-critics, we do not require adaptation for our \textit{loss network} at test time.
Most importantly, compared to the above methods, we propose to learn a loss function that is designed to operate on \textit{the entire dataset} at once, thus leveraging relational information between datapoints.
We achieve this by using a relation network \cite{santoro2017simple} that looks at pairwise combinations of datapoints. As we show in this paper, this leads to a significant improvement in terms of performance.

\section{Regression}
\subsection{Details}

In the sine curve regression task, we follow the architecture used in the original paper for CAVIA \cite{zintgraf2019fast} (a neural network with two hidden layers and 40 nodes each).
Unless otherwise stated, by default we use 5 context parameters. 
In addition, a batch of 25 tasks is used per meta-update. 
We train for 50,000 iterations, with early stopping on a meta-validation set of 100 newly sampled tasks.
During testing, we presented the model with $p$ (default $p=10$) datapoints from 1000 newly sampled tasks and measured MSE over 100 linearly spaced test points. 
In the meta-update step, the task-agnostic parameters of the prediction network is updated using the Adam optimiser with the standard learning rate of $0.001$ which is annealed every 5,000 steps by multiplying it by $0.9$.

To allow a fair comparison, in VIABLE we use the same architecture as CAVIA for the prediction network. 
For both the relation loss network and the simple loss network, we use a neural network with three hidden layers of 32 nodes each. 
In the meta-update step, the parameters of the loss network is learned along with the task-agnostic parameters of the prediction network using the Adam optimiser with the standard learning rate of $0.001$ which is annealed every 5,000 steps by multiplying it by a factor of $0.9$.

Both VIABLE and CAVIA are trained with a single inner-loop gradient step with an inner loop learning rate of 1.0. 

\subsection{Additional Results}
\label{appendix:regression-experiments}
\begin{table}[H]
  \centering
  \begin{tabular}{lccccc}
    \toprule
                            & \multicolumn{5}{c}{Number of Context Parameters}                   \\
    \cmidrule(r){2-6}
    Method                  & 1     & 2     & 3     & 4     & 5    \\
    \midrule
    MAML                    & 0.29 (0.02)      & 0.24 (0.02)      & 0.24 (0.02)      & 0.23 (0.02)      & 0.23 (0.02)          \\
    CAVIA                   & 0.84 (0.06)      & 0.21 (0.02)      & 0.20 (0.02)       & 0.19 (0.02)       & 0.19 (0.02)           \\
    simVIABLE            & 0.75 (0.05)      & 0.14 (0.01)      & 0.15 (0.01)      & 0.14 (0.01)      & 0.16 (0.01)           \\
    relVIABLE            & \textbf{0.57 (0.05)}      & \textbf{0.02 (0.00)}       & \textbf{0.04 (0.00)}      & \textbf{0.03 (0.00)}       & \textbf{0.01 (0.00)}            \\
    
    \bottomrule
  \end{tabular}
  \caption{Results for the sine curve regression task. Shown is the mean-squared-error (MSE) for varying number of context parameters, with 95\% confidence intervals in brackets.}
  \label{table:vary-context-parameters}
\end{table}

\begin{figure}
    \centering
     \includegraphics[width=0.5\linewidth]{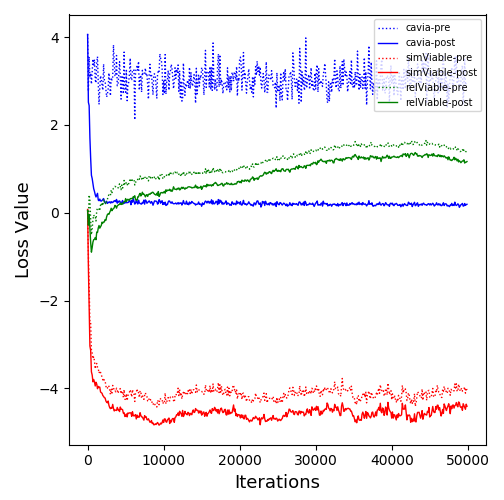}
    \captionof{figure}{Pre-update and post-update test-time loss of each method on the sine curve task. The task specific loss of CAVIA is mean-squared-error (MSE) loss. The task specific loss of VIABLE is the output of the learned loss network.}
    \label{fig:Loss-Value-Plot}
\end{figure}

\section{Classification}

\subsection{Problem Setting}
In $N$-way $K$-shot classification, a task is a random selection of $N$ classes. 
The model gets to see $K$ examples per class from which the model is expected to learn to classify unseen images from the $N$ classes. 
The Mini-Imagenet dataset is divided into training, validation, and test metasets with 64 classes, 16 classes, and 20 classes respectively in which there are 600 images per class.
We use an open-source dataset of Mini-Imagenet embeddings made available by \cite{rusu2019meta}. 
The embeddings are each of size 640. 

\subsection{Model Details}
In CAVIA, our model uses a single hidden layer of size 800 and 100 context parameters.
To ensure fairness, we use the same architecture for the prediction network in VIABLE.
In simVIABLE, our loss network consisted of two hidden layers of 64 nodes each, and in relVIABLE, it consisted of two hidden layers of 1500 nodes each.
Both VIABLE and CAVIA are trained with two inner-loop gradient steps along with an inner-learning rate of 1.0. 
In the meta-update step, VIABLE (prediction network and loss network) and CAVIA are both trained using the Adam optimiser with the standard learning rate of $0.001$ which is also annealed every 5,000 steps by multiplying it by a factor of $0.9$.

\subsection{Further Experiments}

We perform an additional experiment that evaluates CAVIA and VIABLE's ability to generalise to different amount of shots than seen during training. In this experiment, we train on 5-way 5-shot tasks and evaluate on 5-way k-shot where k varies from 1 to 9. Table \ref{table:classification-vary-num-shots} shows both variants of VIABLE significantly outperform CAVIA in generalising at test time to tasks which have a different amount of data than during meta-training. In the case of $k=1$, the relation loss network calculates a loss using the same input in a pair with itself.

\begin{table}
  \centering
  \begin{tabular}{lcccc}
    \toprule
                            & \multicolumn{4}{c}{Number of Shots: 5-way k-shot}                   \\
    \cmidrule(r){2-5}
    Method                  & 1     & 2     & 3     & 4          \\
    \midrule
    CAVIA                   & $50.36 \pm 0.49\%$ & $58.94 \pm 0.46\%$ & $62.64 \pm 0.46\%$ & $65.61 \pm 0.44\%$ \\
    simVIABLE               & $53.94 \pm 0.49\%$ & $62.19 \pm 0.44\%$ & $65.52 \pm 0.43\%$ & $68.16 \pm 0.41\%$  \\
    relVIABLE               & $\mathbf{55.03 \pm 0.48}\%$ & $\mathbf{63.02 \pm 0.44}\%$ & $\mathbf{66.59 \pm 0.42}\%$ & $\mathbf{68.79 \pm 0.42}\%$ \\
    \bottomrule
  \end{tabular}
  
  \begin{tabular}{ccccc}
    \toprule
    \multicolumn{5}{c}{Number of Shots: 5-way k-shot}                   \\
    \cmidrule(r){1-5}
    5                   & 6 & 7 & 8 & 9 \\
    \midrule
    $67.07 \pm 0.45\%$                    & $68.32 \pm 0.43\%$ & $69.13 \pm 0.43\%$ & $70.16 \pm 0.43\%$ & $69.95 \pm 0.44\%$  \\
    $69.32 \pm 0.41\%$        & $70.10 \pm 0.40\%$ & $71.03 \pm 0.40\%$ & $72.01 \pm 0.39\%$ & $71.79 \pm 0.39\%$ \\
    $\mathbf{70.23 \pm 0.41}\%$       & $\mathbf{71.06 \pm 0.39}\%$ & $\mathbf{71.90 \pm 0.40}\%$ & $\mathbf{72.57 \pm 0.39}\%$ & $\mathbf{72.80 \pm 0.39}\%$ \\
    \bottomrule
  \end{tabular}
  \caption{Results for Mini-Imagenet. Shown is the accuracy for 5-way k-shot while being pre-trained on 5-way 5-shot, with 95\% confidence intervals in brackets.}
  \label{table:classification-vary-num-shots}
\end{table}